\documentclass[conference]{IEEEtran}
\IEEEoverridecommandlockouts
\usepackage{cite}
\usepackage{amsmath,amssymb,amsfonts}
\usepackage{algorithmic}
\usepackage{dblfloatfix}
\usepackage{graphicx}
\usepackage{textcomp}
\usepackage{xcolor}
\usepackage[export]{adjustbox}
\def\BibTeX{{\rm B\kern-.05em{\sc i\kern-.025em b}\kern-.08em
    T\kern-.1667em\lower.7ex\hbox{E}\kern-.125emX}}
\begin{document}

\makeatletter
\newcommand{\newlineauthors}{%
  \end{@IEEEauthorhalign}\hfill\mbox{}\par
  \mbox{}\hfill\begin{@IEEEauthorhalign}
}
\makeatother

\title{On adversarial patches: real-world attack on ArcFace-100 face recognition system\\
}


\author{\IEEEauthorblockN{Mikhail Pautov\IEEEauthorrefmark{1},
Grigorii Melnikov\IEEEauthorrefmark{2},
Edgar Kaziakhmedov\IEEEauthorrefmark{3},
Klim Kireev\IEEEauthorrefmark{4},
Aleksandr Petiushko\IEEEauthorrefmark{5}}
\IEEEauthorblockA{
\IEEEauthorrefmark{1}\IEEEauthorrefmark{2}\IEEEauthorrefmark{3}\IEEEauthorrefmark{4}\textit{Skolkovo Institute of Science and Technology}; Moscow, Russia, \\
\IEEEauthorrefmark{1}\IEEEauthorrefmark{2}\IEEEauthorrefmark{3}\IEEEauthorrefmark{4}\IEEEauthorrefmark{5}\textit{Intelligent Systems Lab}; Huawei Moscow Research Center; Moscow, Russia\\
Email: \IEEEauthorrefmark{1}mikhail.pautov@phystech.edu,
\IEEEauthorrefmark{2}grigorii.melnikov@skoltech.ru,
\IEEEauthorrefmark{3}edgar.kaziakhmedov@phystech.edu,\\
\IEEEauthorrefmark{4}klim.kireev@skoltech.ru,
\IEEEauthorrefmark{5}petyushko.alexander1@huawei.com}}

\maketitle

\begin{abstract} 
Recent works showed the vulnerability of image classifiers to adversarial attacks in the digital domain. However, the majority of attacks involve adding small perturbation to an image to fool the classifier. Unfortunately, such procedures can not be used to conduct a real-world attack, where adding an adversarial attribute to the photo is a more practical approach. In this paper, we study the problem of real-world attacks on face recognition systems. We examine security of one of the best public face recognition systems, LResNet100E-IR with ArcFace loss, and propose a simple method to attack it in the physical world. The method suggests creating an adversarial patch that can be printed, added as a face attribute and photographed; the photo of a person with such attribute is then passed to the classifier such that the classifier's recognized class changes from correct to the desired one. Proposed generating procedure allows projecting adversarial patches not only on different areas of the face, such as nose or forehead but also on some wearable accessory, such as eyeglasses.
\end{abstract}


\section{Introduction}

In the last years, much of research was done in the field of attacks on face recognition systems. Nowadays, one of the best face recognition systems  to test different attacking approaches  is LResNet100E-IR with special additive angular margin loss, ArcFace. The network maps an image to a feature vector such that intra-class distance tends to be small and  inter-class distance stays large. In the original paper \cite{deng2019arcface} it is observed that the accuracy in classification problems of this network is comparable to state-of-the-art models. 

Although deep neural networks are efficient in image classification, they are vulnerable to adversarial samples \cite{szegedy2013intriguing}. Majority of prior works has focused on exploiting such vulnerabilities through imperceptible changes of an image, which are constrained by some norm, e.g. $L_1,\ L_2,\ L_\infty$. In such cases, an attacker is trying to fool classifier but at the same time keeps perceptual similarity between original and adversarial images. Later the authors of \cite{brown2017adversarial} succeeded in localizing of image changes. They generated the patch that can be placed anywhere within the field of view of a classifier and managed to conduct an attack in the real world. 

Attack on face recognition neural network is more complex in comparison to attack on just image classifier. It is known for a long time that several areas on a face influence classification more than the others \cite{heisele2004components}. So, the result of classification of the photo with an adversarial patch depends on an area where the patch is located.

Authors of \cite{sharif2016accessorize} investigated the possibility to construct an attack inconspicuous for the human eye. They managed to conduct an attack in the real world with the use of eyeglasses as an adversarial attribute. These glasses were constructed to impersonate a person, but were designed for the face of attacker with a fixed orientation of his head. 

In \cite{komkov2019advhat} the problem of dodging the open face recognition system, ArcFace, was studied. Authors proposed an attack by generating a colored patch that may be placed on a hat and lead to incorrect work of the network in online scenario in physical world.

The goal of this paper is to make a starting point in the improvement of the security of the open face recognition system, ArcFace. Our approach is to explore its vulnerability to adversarial attacks and mislead the network using the proposed method. It should be mentioned that the network and its weights are available on the Internet and the adversarial attack approach is considered as well-known, so the research conducted does not violate any regulations.

The outline of current research is presented below:
\begin{itemize}
\item A simple adversarial patch generation procedure is proposed;
\item The procedure allows to project adversarial patch on different surfaces on a face;
\item Effectiveness of adversarial patches placed on different positions on the face was studied;
\item A real-world attack on one of the best public face recognition systems with the use of this procedure is conducted;
\item The attacking pipeline can be easily implemented since the gained patch is a gray-scale image. 
\end{itemize}

\section{The main concept and related works}
In this section, we describe a concept of adversarial attacks and present an overview of methods and techniques which are often used in the physical domain. 

\subsection{The concept of adversarial attacks}
An adversarial attack on face recognition is a technique to fool some recognition system through a change in input such that the output of the system changes from correct to another one. Although adversarial attack may be conducted in the digital domain (where the input to the classifier may be changed, for example, pixel-wise), it is much harder to construct such an attack in the physical world. However, in \cite{sharif2016accessorize}, \cite{kurakin2016adversarial}, it was shown that it is possible with the efficiency comparable to the one in the digital domain.

There are several possible classifications of adversarial attacks on face recognition systems (i.e. on classifiers).
One way is to determine how the output class should change:
\begin{itemize}
    \item targeted attack -- the adversary changes the output classification of input to the desired one;
    \item untargeted attack -- the adversary leads to misclassification of input.
\end{itemize}{}
In \cite{qiu2019review}, the classification of attacks based on what is known about the neural network is provided. In white-box attacks, parameters of the model, as well as its structure and training procedure, are known. In contrast, in black-box scenarios, none of the above is known.

\subsection{Related works}
In recent years, much research had been done in the field of attacks in the physical domain.

In \cite{Goodfellow2014ExplainingAH}, one of the most straightforward attacking approaches, Fast Gradient Sign Method, FGSM, was proposed. This method refers to generating of adversarial examples through adding to the initial image the perturbation  
\begin{align}
    \eta = \varepsilon sign(\nabla_{x}J(\theta, x, y)), 
\end{align}

where $\theta$ is vector of parameters of the model, $x$ is input to the model, $y$ are the targets associated with $x$ and $J(\theta, x, y)$ is the cost function used to train the neural network. Note that in case of targeted attack, $y$ are desired targets and $\varepsilon < 0$ and in case of untargeted attack, $y$ are incorrect targets and $\varepsilon > 0$.

In \cite{dong2018boosting}, it was discovered that adding the momentum term into the iterative process for attacks lead to more stable optimization trajectory. It is determined that adversarial examples generated with the iterative method with the use of momentum are more suitable for white-box attacks than the ones generated without the use of momentum.

In \cite{Athalye2017SynthesizingRA}, the question of the robustness of adversarial examples under real-world image transformations was studied, and Expectation Over Transformation (EOT) algorithm was proposed.
This algorithm helps to generate an adversarial example taking into account a set of transformations which usually spoils the transferability of an image to the real-world.

Given the distance function $\rho: X \times X \to \mathbb{R}$, distance bound $\varepsilon$, set of transformations $T$ and $y_\text{adv}$ as desired adversarial classes corresponding to $x$, EOT approach may be formulated as  the following optimization problem:

\begin{align}
    X_\text{adv} = \arg\max_{z} \mathbb{E}_{t \in T} \Big[\mathbb{P}(y_\text{adv} | t(x+z))\Big],
\end{align}{}

such that $\mathbb{E}_{t \in T}\big[\rho(x+z, x)\big] < \varepsilon$ and $x + z \in [0,1]^d$. In this formulation $z \in [0,1]^d$ is an additive noise.  This approach helps to imitate such  transformations as camera noise or viewpoint shifts. 

The most similar research to ours was conducted in \cite{komkov2019advhat}. In contrast, there was used different EoT approach as well as patch structure and its position. More than that, we used not colored, but black and white patches.

\section{Methodology}

It should be mentioned that usually face recognition process in the real world consists of two sequential parts. Firstly, the image of a person is passed to face detection neural network, which obtains a bounding box with a face. Secondly, the obtained bounding box is passed to the classifying neural network,  LResNet100E-Ir in our case. Nowadays, there is a variety of face detection neural networks. In our experiments, we use the multi-task cascaded CNN based framework (MTCCN), proposed in \cite{zhang2016joint}. 

In this paper, we propose a pipeline to generate a broad class of adversarial patches, which can fool white-box models and should be fitted for a given person. In this section, we describe the proposed pipeline and give a detailed explanation of each stage. 

The attacking strategy may be explained as a sequence of the following steps: 
\begin{itemize}
\item choose desired patch location and its shape, 
\item print patch with a chessboard pattern and apply it on a face or wearable attribute, 
\item take a photo of your face and mark corners of the chessboard on a photo,
\item repeat steps mentioned before to get photos with different head rotations,
\item project adversarial patch on the chessboard pattern,
\item get facial landmarks using face detector (MTCNN),
\item warp image to standard $112\times112$ ArcFace input using similarity transformation and obtained landmarks,
\item solve the corresponding optimization problem.
\end{itemize}

\subsection{Projective transformation and chessboard pattern}\label{AA}

\subsubsection{Non-linear 2d transformation}
To approximate non-linear 2d transformation, we apply projective transformation separately to each cell of the chessboard grid. Regular grid $G = \{G_i\}$ of pixels $G_i = (x_{i}^{t}, y_{i}^{t})$ forming output map. $(x_{i}^{t}, y_{i}^{t})$ denotes chessboard sticker's coordinates on the photo. By four corner pixels of $j$'th chessboard cell and corresponding four marked pixels of projected cell we can obtain $\mathrm{M}_{\theta_j}$ matrix of spatial transformation, so sampling points of $j$'th cell:  

\begin{align}
\left(\begin{array}{c}{x_{i_j}^{s}} \\ {y_{i_j}^{s}}\end{array}\right)=\mathcal{T}_{\theta_j}\left(G_{i_j}\right)=\mathrm{M}_{\theta_j}\left(\begin{array}{c}{x_{i}^{t}} \\ {y_{i}^{t}} \\ {1}\end{array}\right),
\end{align}
where $(x_i^s, y_i^s)$ are the source coordinates in the input map, they define the spatial location in the adversarial patch where sampling kernel is applied to get the output map. Given the expression above, the target grid for the adversarial patch is obtained as a union of target coordinates of each cell:

\begin{align}
\left(\begin{array}{c}{x_{i}^{s}} \\ 
{y_{i}^{s}}\end{array}\right) = \{ \mathcal{T}_{\theta_j} (G_{i_j}) \}
\end{align}
To speed up the training procedure, we should precalculate and store sampling points. Next step is differentiable image sampling. In more details, this procedure described in \cite{jaderberg2015spatial}. Scheme of transformations is illustrated in \textbf{Fig. \ref{fig:2dtran}}.

\begin{figure}[htbp]
\centerline{\includegraphics[width=11cm]{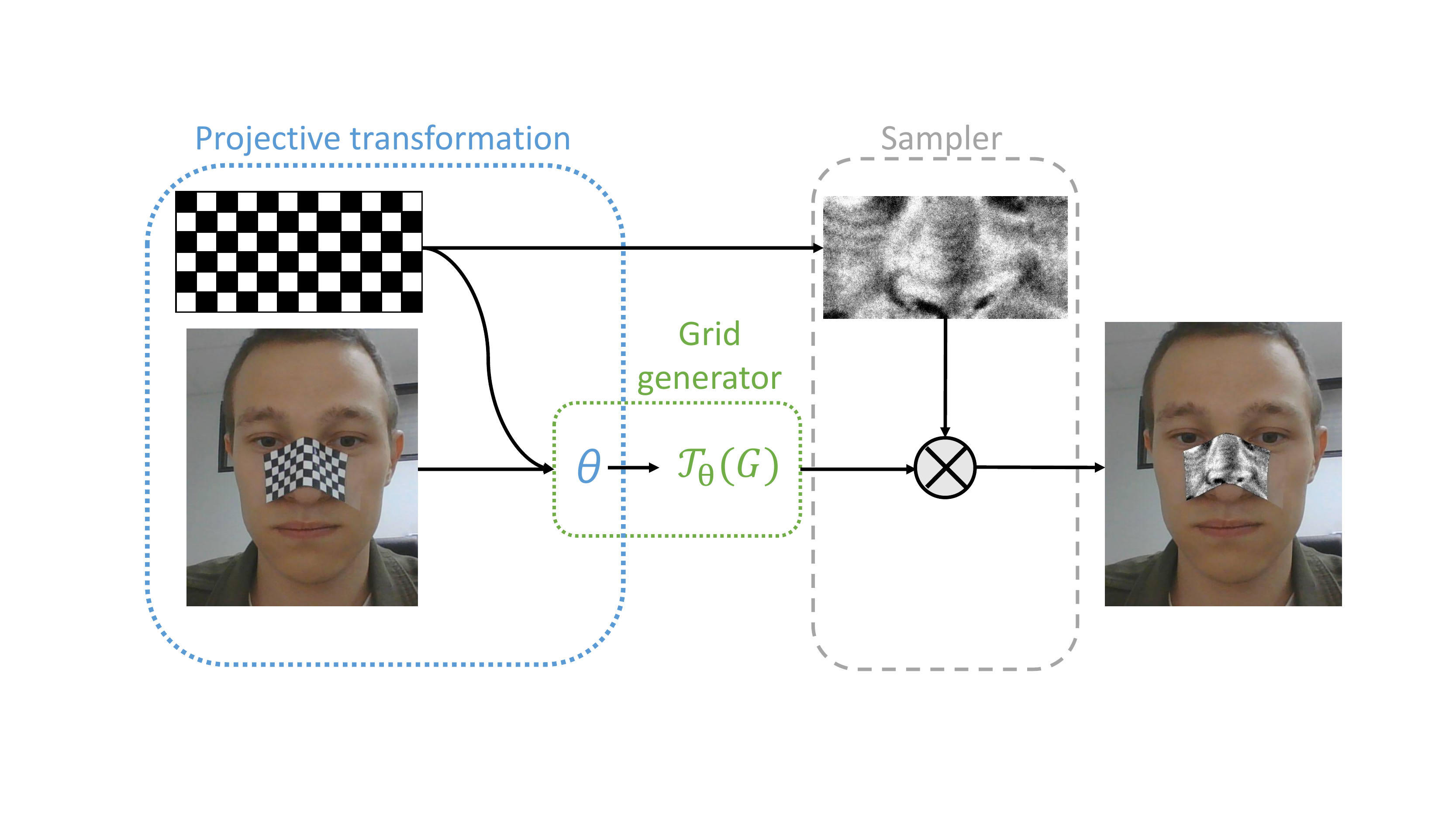}}
\caption{Procedure of adversarial sticker application.}
\label{fig:2dtran}
\end{figure}

\subsubsection{Linear 2d transformation}

In case when the application of adversarial patch could be simulated by a linear transformation (e.g. in case of eyeglasses), we repeat steps mentioned in the previous section just for one cell.

\subsection{Similarity transform and face detector}
MTCNN face detection system finds facial landmarks, which are then used to obtain a matrix of similarity transformation $A_{\phi}$ and sampling points $\mathcal{T_\phi}(G')$ for this warp. 
The main reason why it is necessary to apply sticker before similarity transformation is image sampling. Sampling kernel takes values of neighbouring pixels and copies one of them or return averaged value. Application patch after similarity transformation tends to inconsistent values of pixels on the boundaries of the patch.

\subsection{Loss function and training procedure}
The goal of the attack is to generate adversarial patch which can not only mislead face recognition system but also should be good-looking. We formulate the second objective by restricting neighbouring pixels to have a similar colour. We obtain the adversarial sticker minimizing the following objective function: 
\begin{align}
    \mathcal{L}(X, p) = \mathcal{L}_\text{adv}(X, p) + \tau TV(p),\ p \in [0, 1]^{m\times n} \text{ -- patch,} 
\end{align}{}
where $X$ is a batch of photos of the attacker with different shooting conditions (i.e. set of transforms $T$ described above) and total variation loss, or $TV$ loss, preserve high perceptual quality and ensures that the difference between neighbouring pixels is imperceptible \cite{rudin1992nonlinear}, as far as photos taken from the camera do not have tremendous shift in values of neighbouring pixels.

TV loss is the function of pixel values of a patch $p$ and is defined as follows:

\begin{align}
    TV(p) = \sum_{i,j} \sqrt{\bigl(p_{i,j} - p_{i, j+1}\bigr)^2 + \bigl(p_{i,j} - p_{i+1, j}\bigr)^2}
\end{align}

$\mathcal{L}_\text{adv}$ is the cosine similarity loss that minimizes the similarity between an embedding of the photo with patch and initial embedding of the person. This term depends on type of attack. In case of untargeted attack:
\begin{align}
    \mathcal{L}_\text{adv}(X, p) = \mathbb{E}_{t \in T, x \in X}\Big[-cos(e_{x_{t}}, e)\Big],
\end{align}{}
where $x$ is a photo from batch $X$ corresponding to some shooting condition, $e$ is ground truth embedding of attacker. 

In case of targeted attack:
\begin{align}
    \mathcal{L}_\text{adv}(X, p) = \mathbb{E}_{t \in T, x \in X} \Big[cos(e_{x_{t}}, e_{x'})\Big], 
\end{align}{}
where $e_{x'}$ is an embedding corresponding to desired person or to the closest class except for ground truth class.

Here $e_{x_{t}}$ is an embedding corresponding to the photo $x_t$ of attacker with applied patch. Procedure of patch application may be denoted in a following way. Given a sampler function $A$, photo $x$ and corresponding grid $\mathcal{T}_\theta(G)$, $x_t$ denotes image with applied patch:

\begin{align}
    x_t = A(x,\ t(p),\ \mathcal{T}_\theta(G)).
\end{align}

It should be mentioned that all the operations described above are differentiable. According to \cite{dong2018boosting}, gradients through this iterations  $\nabla_{p} \mathcal{L}(X, p)$ are accumulated. Given decay factor $\mu$,
\begin{align}
    {g} := \mu {g} + \frac{ \nabla_{p}\mathcal{L}(X, p)}{\| \nabla_{p}\mathcal{L}(X, p)\|_{1}}
\end{align}{}

and patch $p$ is updated as follows:

\begin{align}
    p := p - \varepsilon\ sign (g).
\end{align}{}

\begin{figure*}
  \adjincludegraphics[width=\textwidth,trim={0 0 0 4.5cm},clip]{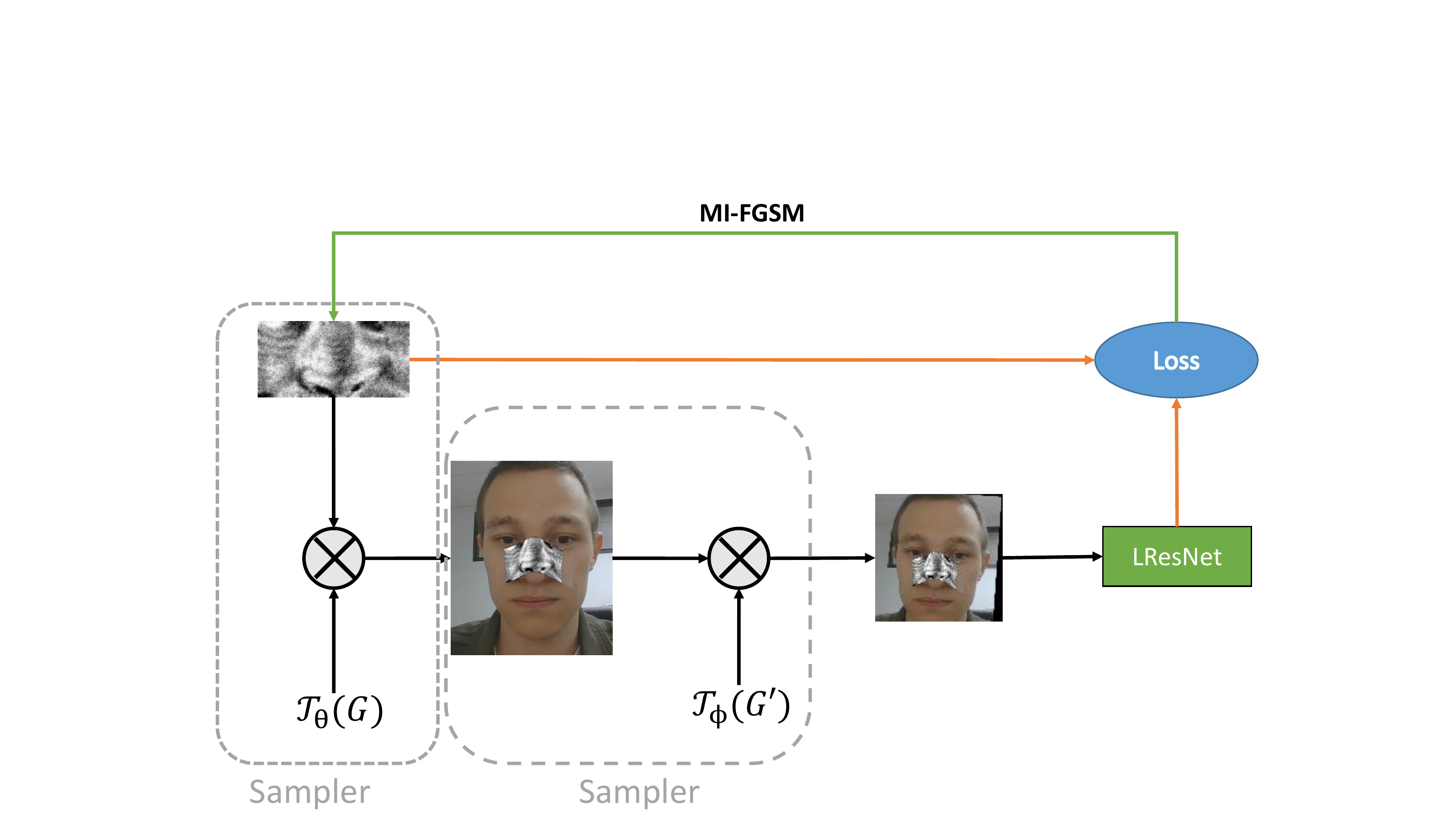}
  \caption{Attacking pipeline}
  \label{fig:pipeline}
\end{figure*}

The whole attacking pipeline is presented in \textbf{Fig \ref{fig:pipeline}}.

\section{Experiments}

In this section, we describe settings of experiments and its technical details.

\subsection{Data preprocessing}
As a part of the experimental setup, we used pretrained LResNet100E-IR to collect embeddings of different people. For this purpose, we use first 200 classes from CASIA-WebFace \cite{yi2014learning} dataset and photos of the $1^{st}$ and the $2^{nd}$ authors as images of attackers. 

\subsection{Setup of experiments}
It should be mentioned that solving the optimization problem described in the previous section implies patch optimization in the whole RGB colour space. However, colour triplets of the patch in digital space differ significantly from the ones of the printed patch because of the printer's narrow colour spectrum. More than that, triplets change one more time when the printed patch is photographed on camera. Due to these issues, we decided to use not a colour but a grayscale patch. 

It was discovered that iterative FGSM method with momentum works good in the digital domain, but gradient step size $\varepsilon$, decay factor $\mu$ and weight $\tau$ of TV loss affect colours transition to physical domain dramatically. It was empirically estimated that values $\varepsilon = \frac{1}{16}$, $\mu = 0.9$ and $\tau = 1e-3$ during the training leads to the best results on test in the real world.

In addition, we want to mention that initial similarity with ground truth class $cos(e_{x_{t}}, e)$ is between $0.65$ and $0.70$ for all training photos in all three scenarios, whereas initial similarity with desired class $cos(e_{x_{t}}, e_{x'})$ in targeted attacks  is always below $0.30$.

\subsection{Adversarial sticker localization}
In this work, we tested the method proposed above to generate patches for three different areas on the face. For the diversity, one of the generated patches is an eyeglass; two others are stickers on nose and forehead.

\subsubsection{Attacking eyeglasses}
In this case, we use a patch of the form of eyeglasses rim (front view), where the frame and bridge of eyeglasses are used for adversarial pattern application. It turned out that the size of these parts is crucial for the success of the attack. Thus, we decided to use an eyeglasses model with a big frame and of size $16.5 \times 6.4$ cm.  Example of the adversarial patch of this form is presented in \textbf{Fig. \ref{fig:eyeglasses}} and  \textbf{Fig. \ref{fig:eyeglasses_patch}}.

\begin{figure}[htbp]
\centerline{\includegraphics[width=4.5cm]{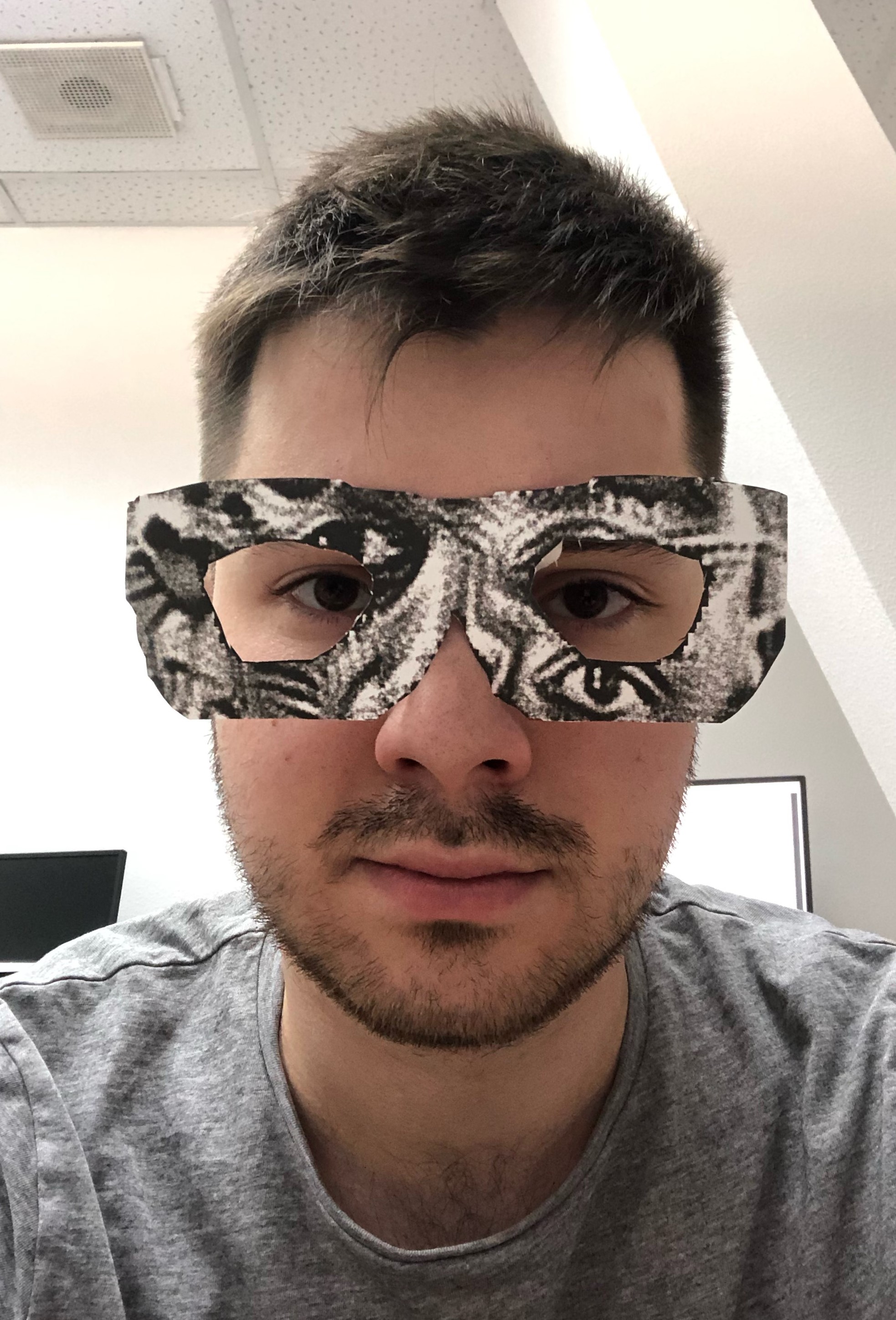}}
\caption{Adversarial eyeglasses}
\label{fig:eyeglasses}
\end{figure}

\begin{figure}[htbp]
\centerline{\includegraphics[width=4.5cm]{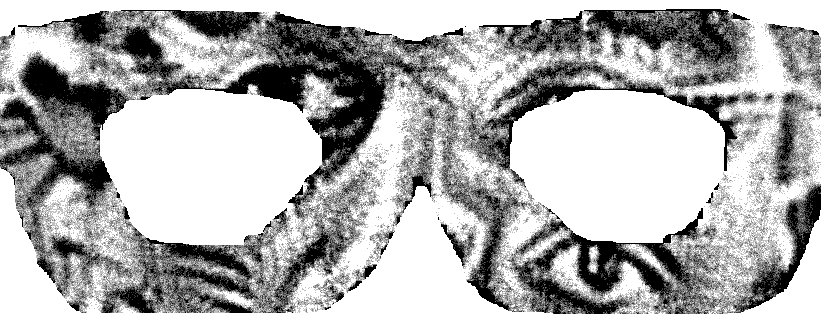}}
\caption{Example of adversarial eyeglasses}
\label{fig:eyeglasses_patch}
\end{figure}

In this scenario, $14$ photos were used for training, $3$ photos were used for validation in the digital domain, and $5$ photos were used for the test in the physical domain.

\subsubsection{Attacking forehead}
In this case, we used a rectangular patch of size $14 \times 5$ cells; the width of one cell is $1.0$ cm. It turned out that the closer patch to eyebrows the better results in a targeted attack. In this scenario, $4$ photos were used for training, $3$ photos were used for validation in digital domain and $3$ photos were used for the test in the physical domain. The patch of this type is illustrated in \textbf{Fig. \ref{fig:forehead}} and in \textbf{Fig. \ref{fig:forehead_patch}}.

\begin{figure}[htbp]
\centerline{\includegraphics[width=4.5cm]{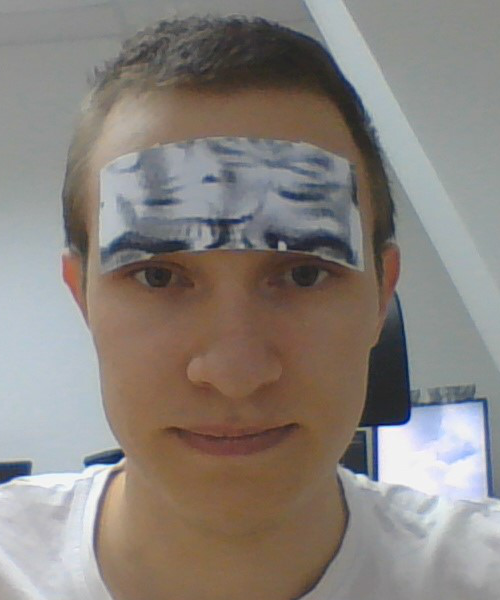}}
\caption{Adversarial forehead}
\label{fig:forehead}
\end{figure}

\begin{figure}[htbp]
\centerline{\includegraphics[width=4.5cm]{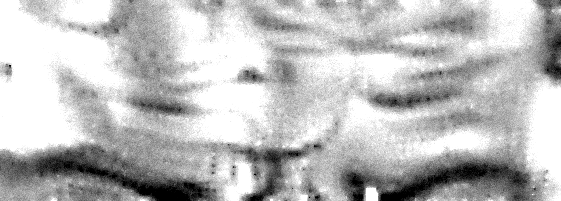}}
\caption{Example of adversarial sticker on forehead}
\label{fig:forehead_patch}
\end{figure}

\subsubsection{Attacking nose}
In this case, we used a rectangular patch of size $12 \times 6$ cells; the width of one cell is $0.7$ cm. An example of adversarial sticker is depicted in \textbf{Fig. \ref{fig:nose}} and \textbf{Fig. \ref{fig:nose_patch}}. We varied length in cells of the patch. During experiments, we found out that the height of the patch less than $6$ cell is not enough for the successful attack even in the digital domain. but it was found out that increasing of size of patch is not enough for transferability  of attack to the physical domain. 

\begin{figure}[htbp]
\centerline{\includegraphics[width=4.5cm]{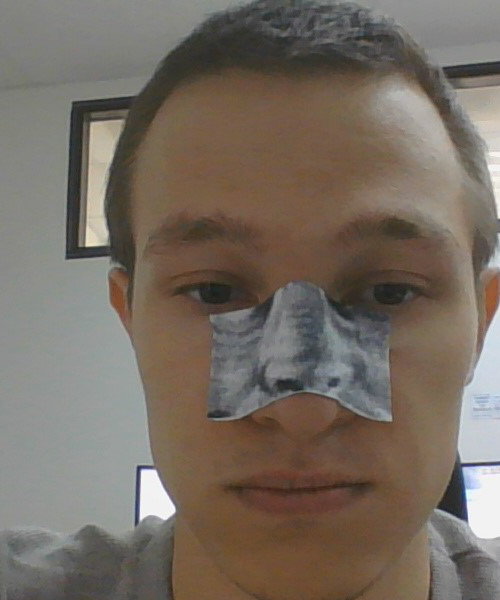}}
\caption{Adversarial nose}
\label{fig:nose}
\end{figure}

\begin{figure}[htbp]
\centerline{\includegraphics[width=4.5cm]{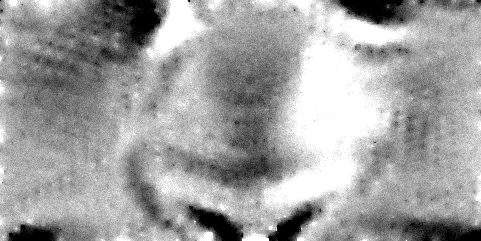}}
\caption{Example of adversarial sticker on nose}
\label{fig:nose_patch}
\end{figure}

\subsection{Results}
In all three above scenarios, training procedure was conducted for $10000$ iterations of the method described before. If on some iteration $\mathcal{L}_\text{adv}(X, p) < 0$, training stops and an obtained patch is printed and tested in the physical world.

Numerical results of the attacks for the first and the second attacking scenarios are presented in Table \ref{tab1}. There we report the mean
similarities $\pm$ standard errors of the mean on training photos $x_{train}$, on validation photos in digital domain $x_{val}$ and on test photos in physical domain $x_{test}$ with both ground truth class and desired class.

\subsection{Discussion}

Numerical experiments showed that it is possible to attack ArcFace in the real world with the use of adversarial stickers placed on eyeglasses or forehead. Although  similarity with ground truth class $cos(e_{x_{test}}, e)$ is just slightly below the similarity with desired class $cos(e_{x_{test}}, e_{x'})$, it is noticeable that neural network can not recognize the attacker as ground truth class. It should be said that the improvement of stability of the proposed technique to different illumination conditions may increase the success of the attack.

In addition, it may be mentioned that one of the interesting outcomes of attack is appearing of sketches of parts of the face on adversarial patches located in corresponding areas. More than that, it is discovered that the position of a patch, as well as its size, dramatically affects the success of attack in the physical domain. In general, the bigger the patch and the closer it to eyes, the better the results of attack in the real world.

\begin{table*}[!tbp]
\begin{center}
\caption{Numerical results of experiments}
 \begin{tabular}{||c c c c c c c c||} 
 \hline
 Patch & Type of attack & $cos(e_{x_{train}}, e)$ & $cos(e_{x_{train}}, e_{x'})$ & $cos(e_{x_{val}}, e)$ & $cos(e_{x_{val}}, e_{x'})$ & $cos(e_{x_{test}}, e)$ & $cos(e_{x_{test}}, e_{x'}) $  \\ [0.5ex] 
 \hline\hline
 Eyeglasses & Targeted & 0.041 $\pm$ 0.052 & 0.648 $\pm$ 0.020 & 0.317 $\pm$ 0.004 & 0.451 $\pm$ 0.021 & 0.305 $\pm$ 0.024 & 0.363 $\pm$ 0.024 \\ 
 \hline
 Forehead & Targeted & -0.053 $\pm$ 0.009 & 0.221 $\pm$ 0.011 &  0.273 $\pm$ 0.007 & 0.421 $\pm$ 0.025  & 0.323 $\pm$ 0.035 & 0.391 $\pm$ 0.021 \\ 
 \hline
\end{tabular}
\label{tab1}
\end{center}
\end{table*}

\section{Conclusion and future work}

We proposed a simple method of creating adversarial patches that can be used to attack state-of-the-art face recognition systems. Our approach was tested in targeted and untargeted attacks on FaceID model LResNet100E-IR, ArcFace@ms1m-refine-v2. In this paper, it was shown that a simple and easily reproducible attacking technique leads to incorrect work of that system not only in the digital domain but also in the physical world. Experiments show that further security research should be conducted to improve the robustness of such networks to attacks of this type. In the future, we focus on construction and utilization of a defense technique to make a network invulnerable to such attacks, especially in the physical world.

\newpage

\bibliographystyle{IEEEtran}
\bibliography{adv}

\begin{thebibliography}{10}
\providecommand{\url}[1]{#1}
\csname url@samestyle\endcsname
\providecommand{\newblock}{\relax}
\providecommand{\bibinfo}[2]{#2}
\providecommand{\BIBentrySTDinterwordspacing}{\spaceskip=0pt\relax}
\providecommand{\BIBentryALTinterwordstretchfactor}{4}
\providecommand{\BIBentryALTinterwordspacing}{\spaceskip=\fontdimen2\font plus
\BIBentryALTinterwordstretchfactor\fontdimen3\font minus
  \fontdimen4\font\relax}
\providecommand{\BIBforeignlanguage}[2]{{%
\expandafter\ifx\csname l@#1\endcsname\relax
\typeout{** WARNING: IEEEtran.bst: No hyphenation pattern has been}%
\typeout{** loaded for the language `#1'. Using the pattern for}%
\typeout{** the default language instead.}%
\else
\language=\csname l@#1\endcsname
\fi
#2}}
\providecommand{\BIBdecl}{\relax}
\BIBdecl

\bibitem{deng2019arcface}
J.~Deng, J.~Guo, N.~Xue, and S.~Zafeiriou, ``Arcface: Additive angular margin
  loss for deep face recognition,'' in \emph{Proceedings of the IEEE Conference
  on Computer Vision and Pattern Recognition}, 2019, pp. 4690--4699.

\bibitem{szegedy2013intriguing}
C.~Szegedy, W.~Zaremba, I.~Sutskever, J.~Bruna, D.~Erhan, I.~Goodfellow, and
  R.~Fergus, ``Intriguing properties of neural networks,'' \emph{arXiv preprint
  arXiv:1312.6199}, 2013.

\bibitem{brown2017adversarial}
T.~B. Brown, D.~Man{\'e}, A.~Roy, M.~Abadi, and J.~Gilmer, ``Adversarial
  patch,'' \emph{arXiv preprint arXiv:1712.09665}, 2017.

\bibitem{heisele2004components}
B.~Heisele and T.~Koshizen, ``Components for face recognition,'' in \emph{Sixth
  IEEE International Conference on Automatic Face and Gesture Recognition,
  2004. Proceedings.}\hskip 1em plus 0.5em minus 0.4em\relax IEEE, 2004, pp.
  153--158.

\bibitem{sharif2016accessorize}
M.~Sharif, S.~Bhagavatula, L.~Bauer, and M.~K. Reiter, ``Accessorize to a
  crime: Real and stealthy attacks on state-of-the-art face recognition,'' in
  \emph{Proceedings of the 2016 ACM SIGSAC Conference on Computer and
  Communications Security}.\hskip 1em plus 0.5em minus 0.4em\relax ACM, 2016,
  pp. 1528--1540.

\bibitem{komkov2019advhat}
S.~Komkov and A.~Petiushko, ``Advhat: Real-world adversarial attack on arcface
  face id system,'' \emph{arXiv preprint arXiv:1908.08705}, 2019.

\bibitem{kurakin2016adversarial}
A.~Kurakin, I.~Goodfellow, and S.~Bengio, ``Adversarial examples in the
  physical world,'' \emph{arXiv preprint arXiv:1607.02533}, 2016.

\bibitem{qiu2019review}
S.~Qiu, Q.~Liu, S.~Zhou, and C.~Wu, ``Review of artificial intelligence
  adversarial attack and defense technologies,'' \emph{Applied Sciences},
  vol.~9, no.~5, p. 909, 2019.

\bibitem{Goodfellow2014ExplainingAH}
I.~J. Goodfellow, J.~Shlens, and C.~Szegedy, ``Explaining and harnessing
  adversarial examples,'' \emph{CoRR}, vol. abs/1412.6572, 2014.

\bibitem{dong2018boosting}
Y.~Dong, F.~Liao, T.~Pang, H.~Su, J.~Zhu, X.~Hu, and J.~Li, ``Boosting
  adversarial attacks with momentum,'' in \emph{Proceedings of the IEEE
  conference on computer vision and pattern recognition}, 2018, pp. 9185--9193.

\bibitem{Athalye2017SynthesizingRA}
A.~Athalye, L.~Engstrom, A.~Ilyas, and K.~Kwok, ``Synthesizing robust
  adversarial examples,'' in \emph{ICML}, 2017.

\bibitem{zhang2016joint}
K.~Zhang, Z.~Zhang, Z.~Li, and Y.~Qiao, ``Joint face detection and alignment
  using multitask cascaded convolutional networks,'' \emph{IEEE Signal
  Processing Letters}, vol.~23, no.~10, pp. 1499--1503, 2016.

\bibitem{jaderberg2015spatial}
M.~Jaderberg, K.~Simonyan, A.~Zisserman \emph{et~al.}, ``Spatial transformer
  networks,'' in \emph{Advances in neural information processing systems},
  2015, pp. 2017--2025.

\bibitem{rudin1992nonlinear}
L.~I. Rudin, S.~Osher, and E.~Fatemi, ``Nonlinear total variation based noise
  removal algorithms,'' \emph{Physica D: nonlinear phenomena}, vol.~60, no.
  1-4, pp. 259--268, 1992.

\bibitem{yi2014learning}
D.~Yi, Z.~Lei, S.~Liao, and S.~Z. Li, ``Learning face representation from
  scratch,'' \emph{arXiv preprint arXiv:1411.7923}, 2014.

\end{thebibliography}


\end{document}